\documentclass[conference]{IEEEtran}
\IEEEoverridecommandlockouts
\usepackage{multirow}
\usepackage{multicol}
\usepackage{natbib}
\usepackage{cite}
\usepackage{amsmath,amssymb,amsfonts}
\usepackage{algorithmic}
\usepackage{graphicx}
\usepackage{textcomp}
\usepackage{xcolor}
\def\BibTeX{{\rm B\kern-.05em{\sc i\kern-.025em b}\kern-.08em
    T\kern-.1667em\lower.7ex\hbox{E}\kern-.125emX}}
\begin{document}

\title{Advances in Multiple Instance Learning for Whole Slide Image Analysis: Techniques, Challenges, and Future Directions\\
{\footnotesize \textsuperscript{}}

}

\author{\IEEEauthorblockN{Jun Wang}
\IEEEauthorblockA{
jwang699-c@my.cityu.edu.hk}
\and
\IEEEauthorblockN{Yu Mao}
\IEEEauthorblockA{
 yumao7-c@my.cityu.edu.hk}
\and
\IEEEauthorblockN{Nan Guan}
\IEEEauthorblockA{
nanguan@cityu.edu.hk}
\and
\IEEEauthorblockN{Chun Jason Xue}
\IEEEauthorblockA{
jason.xue@mbzuai.ac.ae}
}

\maketitle

\begin{abstract}
Whole slide images (WSIs) are gigapixel-scale digital images of H\&E-stained tissue samples widely used in pathology. The substantial size and complexity of WSIs pose unique analytical challenges. Multiple Instance Learning (MIL) has emerged as a powerful approach for addressing these challenges, particularly in cancer classification and detection. This survey provides a comprehensive overview of the challenges and methodologies associated with applying MIL to WSI analysis, including attention mechanisms, pseudo-labeling, transformers, pooling functions, and graph neural networks. Additionally, it explores the potential of MIL in discovering cancer cell morphology, constructing interpretable machine learning models, and quantifying cancer grading. By summarizing the current challenges, methodologies, and potential applications of MIL in WSI analysis, this survey aims to inform researchers about the state of the field and inspire future research directions.
\end{abstract}


\section{Introduction}
Cancer is a global public health challenge and the leading cause of mortality worldwide, responsible for nearly 10 million deaths in 2020 \citep{sung2021global}. It is a highly complex disease that involves a series of microscopic and macroscopic alterations, with underlying mechanisms and interactions that are not yet fully understood. Machine learning techniques have significantly impacted the field of medical imaging, as they are capable of learning highly complex representations from whole slide images(WSIs)~\citep{song2023artificial,saltz2018spatial}. The transition towards the digitization and automation of clinical pathology, often referred to as computational pathology (CPath), has the potential to enhance diagnostic and prognostic accuracy,  improve the discovery of novel biomarkers, and improve predictions regarding treatment responses~\citep{van2021deep,cifci2023ai,echle2021deep}. Histopathological tasks, such as cancer classification, grading, and subtyping, are crucial to studying cancer diagnosis and progression and developing targeted therapies, as cancer effects and gene expressions can be examined at the cellular and tissue levels in WSIs~\citep{shmatko2022artificial,perez2024guide,murchan2021deep}. 

Despite these notable advances, CPath still present great potential in pathology research and precision medicine. These challenges primarily from the large size of the images and the inherent complexity and variability of the histopathological features they contain.

Traditional machine learning methods through supervised learning for WSI analysis reformulate slide classification as a plaque-level supervised learning classification task, relying on the annotation of region of interst. However, these approaches have some drawbacks. The large amounts of annotated training data is be expensive and time-consuming to obtain, and the inability to detect novel or rare structures that were not present in the training data, since CNN might only learn from one of the multiple discriminative patterns in WSIs. In addition, there are limitations in hardware resource, dut ot the huge computations required. Solutions, such as a downsampling strategy , can solve this but consequently result in a significant loss of discriminative details.

Therefore, weakly supervised learning, such as multiple instance learning, has been gaining popularity in recent years for WSI analysis. WSI classification can be defined as a MIL problem, where a single supervised label is provided for the set of patches that constitute the WSI, and only a subset of patches are assumed to correspond to that label~\citep{song2023artificial}. MIL learning maps a set of patches to labels in three steps(as shown in~\ref{fig:whole}): first, a feature extractor extracts a low-dimensional embedding for each patch; second, an aggregator pools the patch embeddings to form a WSI representation; and third, a predictor is used to map the representation to a WSI label~\citep{van2021deep,perez2024guide}. MIL differs from patch-level learning in that WSI-level labels are no longer assigned to patches, but to the set of patches that constitute the WSI.

In recent years, the application of MIL for WSI analysis has been widely used across various oncological fields~\citep{campanella2019clinical,schirris2022deepsmile,lu2021data,shao2021transmil,naik2020deep,li2021dual,zhang2022dtfd,carmichael2022incorporating,anand2021weakly,qu2022dgmil}. MIL has demonstrated great performance in many histopathological tasks, including tasks for automation and tasks for discovery. Automation histopathological tasks focus on tumor detection~\citep{qiu2021attention,del2022constrained,butke2021end,wang2019rmdl,del2021attention}, cancer subtype classification~\citep{schirris2022deepsmile,hashimoto2022subtype,zhao2021lung}, and tumor grading~\citep{myronenko2021accounting}. (\citep{sandarenu2022survival,yao2020whole}). Histopathological tasks for discovery focus on the prediction of mutation and molecular marker status, predict activation of specific genes, survival rate and response to treatment.

The aim of this survey is to provide a comprehensive overview of widely employed techniques in the field of multiple instance learning approaches, their histopathological tasks, recent advances in improving the model performance, remaining challenges and future potential. The following outline has been prepared to reflect the overall structure of this survey.

\subsection{Motivation and Contributions:}
In this work, we review papers using multiple instance learning on histopathology, where machine learning techniques have been applied to a wide variety of diagnosis, classification, prediction, and prognosis tasks. Only papers that become available online after 2018 are included.

Understanding the direction in which the computational pathology will converge regarding MIL frameworks is crucial. Therefore, this review aims to introduce the major MIL technical developments for WSI classification across various disease. Addtionally, we outline promising research areas focusing on training robust and interpretable models for WSI analysis from diverse tasks of diseases, clinical value, large-scale datasets, and future developments.

Overall, the main contributions of our survey are summarized as follows:

\begin{enumerate}[i.]

    \item providing a comprehensive overview of multiple instance learning approach applied to whole slide images for pathology diagnosis, including the state-of-the-art models in CPath and the framework of WSI analysis using multiple instance learning approaches

    \item categorizing the papers based on machine learning methods
    
    \item summarizing the progress in adopting multiple instance learning into different histopathological tasks and special diseases

    \item discussing the potential of multiple instance learning
    
\end{enumerate}

\section{Multiple Instance Learning in WSI Image Analysis}

In the binary and classic machine learning problem, the model is trained to predict the label, $ y \in \{ 0,1 \}$, for a given input sample $ x \in \mathrm {\mathbb {R}}^ D $. For Multiple instance learning models in whole slide images analysis, the original input WSI $X$ will be split into non-overlapping image patches $X = \{x_{1}, x_{2}, \cdots, x_{K} \} $, which is also called bag of instances. There is a single binary label $Y\in \{ 0,1 \}$ associating with the bag of instances. According to the original MIL assumptions ~\citep{dietterich1997solving}, the MIL problem can be stated in the following form:

\begin{equation}
Y =
    \begin{cases}
        0,& \text{if  $\sum_{k}^{} y_{k} = 0 $ } \\
        1,& \text{otherwise.}
    \end{cases}
\end{equation}

In the MIL problem, the model is composed of the following three-step transformations:
\begin{equation}
S(X) = g( \sigma(f(x_{1}),f(x_{2}),\cdots,f(x_{K})) )
\end{equation}

The function $f$, $g$, and permutation invariant pooling function $\sigma$ determine how the models learn the label probability.

\begin{figure*}[]
    \centering
    \includegraphics[width=\textwidth]{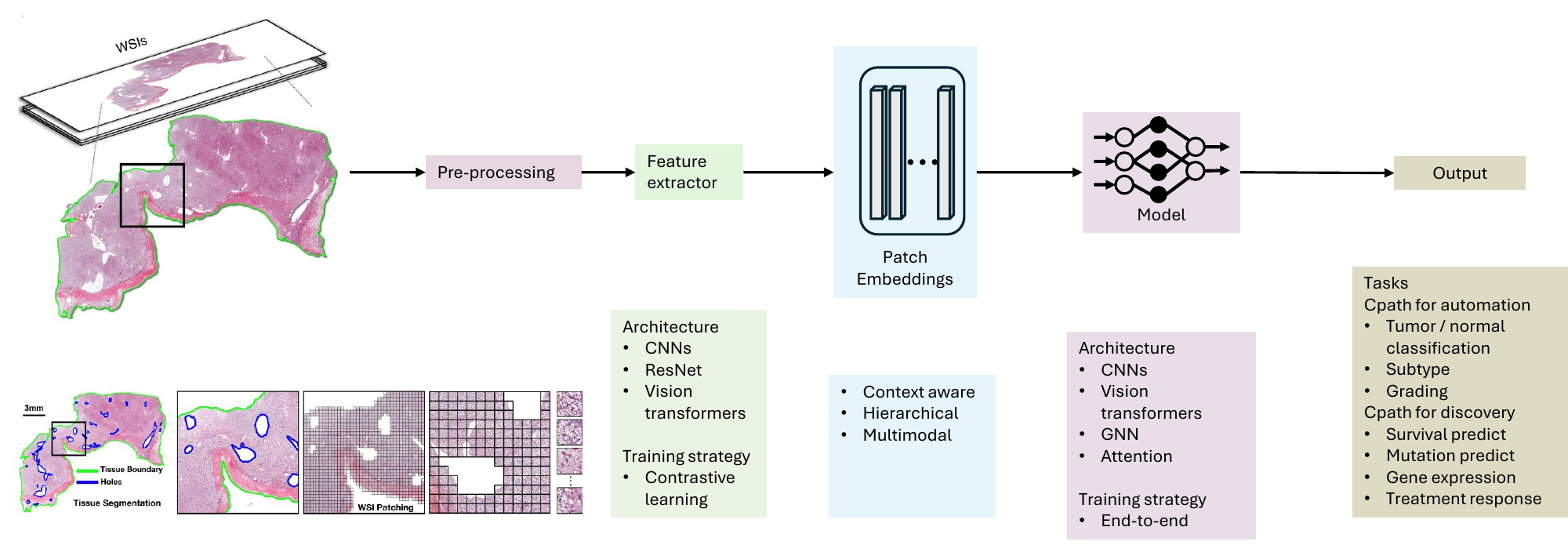}
    \caption{General workflow and tasks in Cpath. Original WSIs and pre-processing is from ~\citep{lu2021data}}
    \label{fig:whole}

\end{figure*}

MIL approaches in Computational pathology can be summarized to the outline as depicted inFig~\ref{fig:whole}. All such methods begin by taking a set of patches extracted from WSIs and learn to map them to WSI-level clinical labels. The original WSI (which is called bag in MIL pipeline) extracts non-overlapping patches firstly, then each patch is individually processed. After pre-processing, each remaining patch is firstly fed into feature extractor(Fig 1.b) obtaining patch embeddings(Fig 1.c), followed by an aggregation(Fig 1.d) forming a WSI representation and finally outputs a bag-level prediction(Fig 1.e).

The following outline has been prepared to reflect the overall structure of this survey.
we summarize this survey by introducing the methodologies about the novel design in structure and the improvement in performance, diverse histopathological tasks and potential of MIL in CPath. Section 2 introduces the deep-learning workflow for WSIs analysis. Section 3 discuss the ehancements in feature extraction. Section 4 explore the improvement and novel designs in embedding level.
The aim of section 5 is to introduce widely used MIL technique applied in computational histopathology from a methodological perspective. Section 6 introduce how these methods are translated into biological contexts in histopathological tasks. Section 7 discusses the potential of multiple instance learning.

\textbf{Pre-Processing.} Before applying AI algorithms, WSIs (Whole Slide Images) undergo tissue segmentation to eliminate background areas using either traditional image processing or deep learning methods. Due to their gigapixel size, direct processing of WSIs is computationally intensive. To manage this, WSIs are typically divided into smaller segments called patches, allowing for individual processing by neural networks within Computational Pathology (CPath) frameworks. The results from these patches can then be aggregated for slide-level or patient-level analysis.

\textbf{Feature Extractor.} Due to memory constraints, the entire set of Whole Slide Image patches cannot be stored simultaneously in GPU memory, making it challenging for Multiple Instance Learning (MIL) to learn feature extractors and predictors jointly. A solution is pre-training the feature extractor on auxiliary tasks like natural image datasets or histopathology images with self-supervised learning to compress patch embeddings. However, models trained on natural images may not be optimal for medical images due to different data distributions. Research suggests using contrastive learning approaches such as SimCLR for better feature extraction in MIL frameworks tailored for histopathology, showing potential improvements over traditional methods like ImageNet-trained models. Some studies also explore integrating contrastive learning into MIL frameworks directly but indicate mixed results in performance compared to other weakly supervised approaches while highlighting possibilities in training ML models with unlabeled datasets.

\textbf{Patch Embedding. }Deep learning models in histopathology divide whole slide images (WSIs) into patches, analyzing them at different magnifications to capture varying levels of detail. Low-magnification patches provide context but lack finer details, while high-magnification patches show more detail but lose contextual information and increase training costs. Pathologists often need to examine WSIs at multiple magnifications for accurate diagnosis, such as using the Gleason grading system for prostate cancer. To address limitations in capturing long-range tissue architecture context, Multiple Instance Learning (MIL) methods with neural networks model interactions between patch embeddings through structures like graphs or sequences. Additionally, multiscale WSI representations and multimodal data fusion enhance decision-making by integrating diverse data sources and modalities, improving AI-driven predictions' robustness and accuracy in medical diagnoses~\citep{lipkova2022artificial}.

\textbf{Aggregation Model. }The aggregation model is utilized to conduct WSI analysis upon features extracted by the Feature extractor. Some works enhance interpretability and accuracy in deep learning applications for histopathology by employing attention-based methods, Vision Transformers (VIT), and Graph Neural Networks (GNN). These techniques improve model transparency by pinpointing critical regions that influence predictions. To boost accuracy, the model addresses challenges such as overfitting caused by large Whole Slide Images (WSI) and limited annotations. It recommends strategies like pseudo labeling and student-teacher distillation within a Multiple Instance Learning (MIL) framework. 

\textbf{CPath Tasks. }In clinical pathology automation tasks, deep-learning methods are being developed to detect tumors, subtype them, and grade them based on morphological features observed in histopathology images. In computational pathology research tasks, these methods investigate the connections between tumor genomic variations and visible morphology in H\&E slides to predict mutational status or other genetic markers crucial for treatment decisions.

\textbf{Clinical impact of CPath}
Multiple instance learning approaches is not only adopted in various automation tasks.(as shown in Fig.\ref{fig:whole})

It also demonstrate the capability of deep learning systems to accurately predict the mutational status of individual genes, deficiencies in DNA repair mechanisms, and overall tumor mutational burdens directly from H\&E slides, and treatment response across various cancer types.

\section{Feature Extractor}
\label{methods}

\begin{figure*}[]
    \centering
    \includegraphics[width=\textwidth]{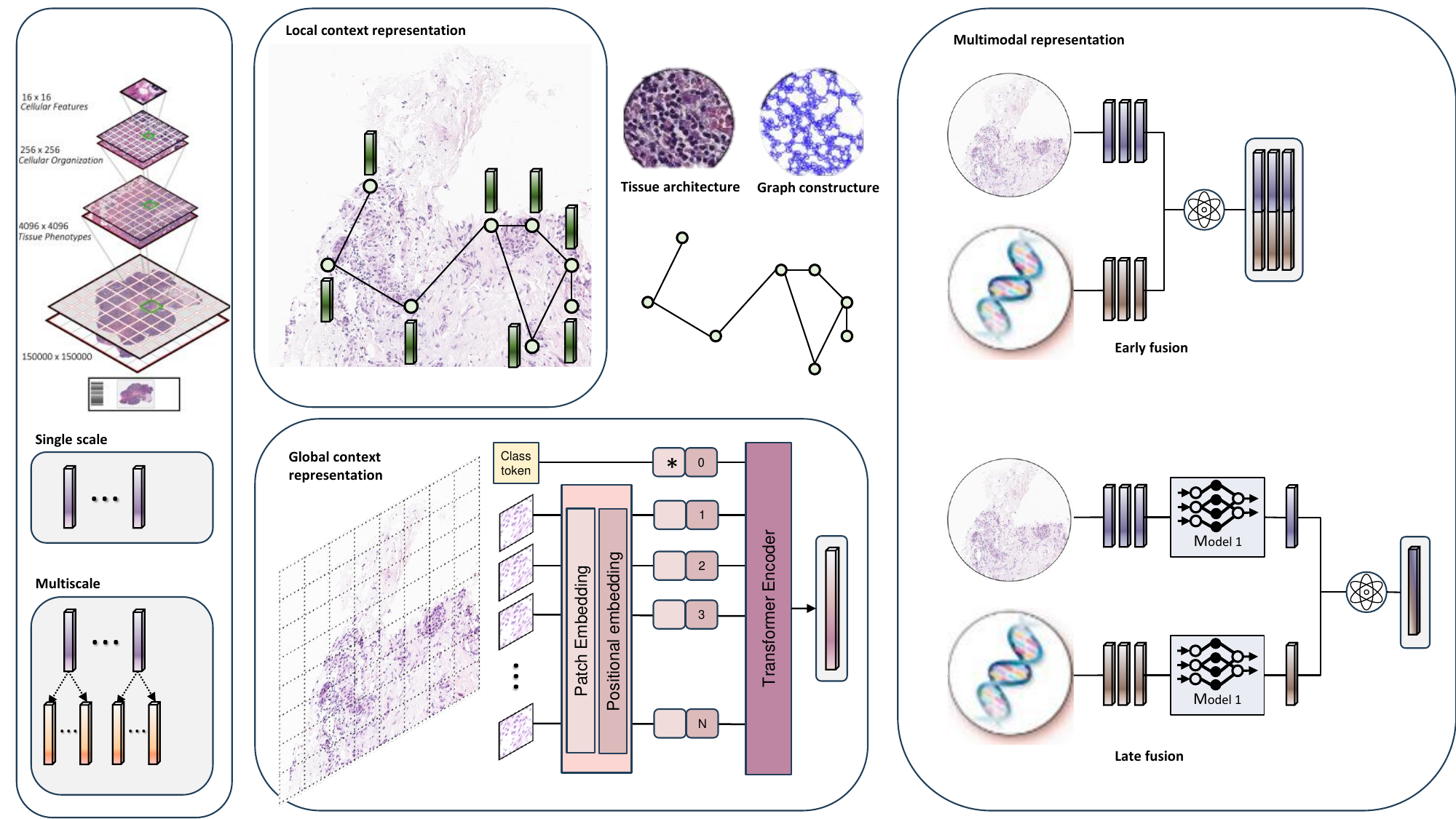}
    \caption{Features in Cpath.~\citep{hipt,lipkova2022artificial} }
    \label{fig:feature}
\end{figure*}
\citep{sandarenu2022survival} introduce neural image compression \citep{tellez2019neural} as the feature extractor into Attention-based MIL framework for survival prediction of triple negative breast cancer.
\citep{yao2020whole} propose a DeepAttnMISL for cancer survival prediction. They utilize VGG-16 pre-trained by ImageNet as the feature extractor and then adopt K-means algorithm to cluster patches based on the extracted feature. These patches are grouped into distinct phenotypic clusters and fed into the Multiple Instance Fully Convolutional Network (MI-FCN) to generate local representations. Attention-based MIL pooling(with Eq.\ref{equation:attention}) to patient-level representations for predictions further aggregates the local representations. 

\citep{lu2019semi} firstly introduce contrastive predictive coding \citep{oord2018representation} into a two-stage weakly supervised framework to prevent the overfitting problem. Before the second stage of the Attention-based MIL framework, contrastive predictive coding is utilized in the first stage, which is a feature network, by pretraining with unlabeled instances to enable the model to learn rich and high-level feature representations. The authors evaluate their model and compare the learned feature via contrastive predictive coding with the extracted feature via ResNet50 pre-trained by ImageNet. Their experiment demonstrates that CPC feature under the MIL framework can significantly and efficiently improve model performance and scale to work on larger WSIs.

\citep{schirris2022deepsmile} demonstrate that a histology-specific pre-trained model, such as SimCLR, extracts more meaningful features compared to ImageNet pre-trained model. They provide experimental evidence that the model can get satisfactory improvement by utilizing feature extractors based on contrastive learning instead of ImageNet pre-trained feature extractors.

\citep{tavolara2022contrastive} also utilize SimCLR as their feature extractor in their MIL framework named SS-MIL. Moreover, they first fuse contrastive learning into the multiple instance learning framework. For each WSI, patches will be embedded into a collection of instances. Each collection will be divided into several subsets, used as MIL bags from one same WSI, to create positive and negative bags for different WSIs. Their experiments don't achieve better performance in histopathological tasks than other weakly supervised MIL approaches but show the potential of combining different datasets to train ML models in histopathological areas. The SS-MIL approach can benefit researchers to train ML models with no-label datasets.

\section{Patch Embedding}

\subsection{Multimodal Methods}
\label{multimodal}

\citep{chen2022pan} employ multimodal MIL to integrate histological images with genomic data for survival prediction. Their approach utilizes multimodal deep learning to jointly analyze pathology images and molecular profiling data for several cancers. Their multimodal MIL algorithm is capable of fusing these diverse modalities to predict outcomes and discover prognostic signatures. 

\subsection{Multi-scale Features}

For cancer subtype classification, one of the difficulties in the clinical diagnosisis for some cancers require pathologists to frequently adjust the magnification of WSIs to identify different tissue phenotypes in different scales. However, previous feature extraction approach cannot achieve this goal with satisfactory performance. Moreover, pathologists often first locate regions containing tumors before performing subtype classification based on identified tumor phenotypes. Additionally, the WSIs from different institutions have different staining conditions, resulting in inaccuracy in classifications. \citep{hashimoto2020multi} incorporate multi-scale mechanism and domain adversarial normalization into the Attention-based MIL framework for cancer subtype to overcome these difficulties in classification. The multi-scale mechanism is implemented to learn distinct global and local features at various scales, thereby enabling the capture of global tissue structure on low scales, and detailed tumor phenotypes on high scales. Domain adversarial normalization is employed to mitigate the impact of different staining conditions and protocols. In their MS-DA-MIL framework, the first stage is to train the feature extractor by attention mechanism
and domain adversarial regularization to generate feature vectors from $S$ scales. After obtaining the feature vectors from the first stage, these feature vectors are subsequently fed into the second stage and then aggregated for bag-level predictions by Attention-based MIL pooling.

\citep{li2021multi} address the challenge of scale variability in the analysis of WSIs for prostate cancer by proposing a multi-resolution multiple instance learning model. For prostate cancer, the Gleason Grading System is the golden standard for diagnosis and prognosis. In current clinical practice, suspicious regions are initially identified using low magnification, and then identified areas are examined at high magnification for Gleason grading. In \citep{li2021multi}, the model is a two-stage Attention-Based MIL architecture with multi-resolution feature extract mechanism. The first stage utilizes the attention network at low magnification to generate a saliency map. This map highlights the significance of each patch in determining the slide-level labels, aiding in the identification of benign and malignant slides. The k-means algorithm is used for clustering. The second stage utilizes the tiles selected from each k-means cluster at high magnification for prostate cancer grade classification. 
\citep{jiang2024transformer,hashimoto2020multi}

\section{Aggregation Model}
\subsection{Ehancements in Architecture}
\subsubsection{Attention mechanism}

In recent years, the attention mechanism plays an important role in medical image analysis. The challenges of MIL have been mentioned in the second part of this survey. Therefore, to overcome these limitations, the attention mechanism is one of the popular solutions that have been proposed.

The attention mechanism in computer vision is motivated by human vision. Humans can easily and efficiently select important regions and certain aspects of their environment. Inspired by human vision, attention mechanisms are developed to focus on the regions that are most relevant for the task, such as regions that contain cancerous cells in a whole slide image. This allows the model to effectively process large, high-resolution images and make more accurate predictions.

In this part, different attention mechanism methods are surveyed and summarized. The contributions of this part are summarized as follows:
\begin{enumerate}[i.]
    \item introduces the foundational backgrounds of attention mechanisms.
    
    \item categories attention mechanisms in multiple instance learning architecture while analyzing whole slide images.

    \item summarizes technical achievements by attention mechanism methods in whole slide image analysis.

\end{enumerate}

\subsubsection{Attention-Based MIL approach}
\citep{ilse2018attention} firstly propose an attention-based multiple instance learning approach. They illustrate that the MIL problem can be a procedure of learning the Bernoulli distribution of the bag label. All the transformations (the transformations of instances, permutation-invariant aggregation function, and transformation to the bag probability) can be parameterized by neural networks. Furthermore, they propose a permutation-invariant aggregation operator, which is called Attention-based MIL pooling, to modify the embedding-level approach. Let $H = \{h_{1}, h_{2}, \cdots, h_{K} \} $ be the embedding result from transformation $f$ for bag $X$ of $K$ instance. The permutation invariant Attention-Based MIL pooling function $\sigma$ is:

\begin{equation}
z = \sum_{k = 1}^{K} a_{k}h_{k}
\label{equation:z}
\end{equation}

\begin{equation}
a_{k} = \frac{ \exp { \{ w^{T}( \tanh  ( V h_{k}^{T}  )\} } }
{\sum_{j = 1}^{K} { \exp { \{ w^{T}( \tanh  ( V h_{j}^{T}  ) \} } } }\qquad
\label{equation:attention}
\end{equation}
where $a_{k} $ is the attention score for $ k^{th}$ instance, and $w$, $V$ are learnable parameters for the attention mechanism proposed by \citep{ilse2018attention}. 
Moreover, they propose a gated attention mechanism with an additional fully connected layer with a sigmoid function to learn more complex relations among instances:
\begin{equation}
a_{k} = \frac{ \exp { \{ w^{T}( \tanh  ( V h_{k}^{T}  ) \odot  sigm ( U h_{k}^{T} )) \}} }
{\sum_{j = 1}^{K} { \exp { \{ w^{T}( \tanh  ( V h_{j}^{T}  ) \odot  sigm ( U h_{j}^{T} )) \}} } }\qquad
\label{equation:gated-attention}
\end{equation}
where $U$ are learnable parameters, and $\odot $ is the element-wise multiplication for the gated attention mechanism. This attention score $a_{k} $ is generated by learning relations among instances within a bag, representing the instances contributing more to the model.

\paragraph{Apply Attention-Based MIL approach directly to different diseases}
Attention-based MIL approach proposed by Ilse et al.(2018) provides a novel insight and significantly impacts the field of whole slide image analysis. The attention-based MIL approach achieves significantly impressive results and is widely used in the diagnosis of different types of cancer. As shown in Table \ref{table:ABMIL-disease}), a large number of studies apply Attention-based MIL \citep{ilse2018attention} directly to their histopathological tasks and clinical problems. The desirable properties of attention-based MIL, such as the interpretability of key instances, make the whole community investigate the potential of Attention-based MIL approaches.

\begin{table*}[!t]
\caption{Summary of papers apply Attention-Based MIL approach to different diseases.}\label{table:ABMIL-disease}
	\resizebox{\textwidth}{!}{
\centering
\begin{tabular}{ p{3cm} | p{4cm} | p{5cm}| p{5cm}}
    \hline		
		Papers& Cancer Types & Histopathological tasks   & Result \\
    \hline
\citep{li2021multi} & prostate cancer& Gleason grading &Acc = 0.927, Auc = 0.982 \\
  \hline
\citep{zhang2021joint} & prostate cancer & Gleason grading & Acc = 0.91, Auc = 0.98 \\
   \hline
\citep{zeng2022artificial} & hepatocellular carcinoma  & activation of specific gene  & Acc = 0.81-0.92, Auc = 0.78-0.91\\
     \hline
\citep{qiu2021attention} & Thyroid cancer & cancer classification & Acc = 0.932\\
     \hline
\citep{anand2021weakly} & Thyroid cancer & activation of specific gene & Auc = 0.96 in TH-TMA17 \\
     \hline
\citep{del2022constrained} & Ulcerative colitis  &  Detect neutrophils and classify into histological remission or adverse outcome & Acc = 0.96, Auc = 0.959 \\
     \hline
\citep{carmichael2022incorporating} & Ulcerative colitis  & Survival Prediction   & 1\% higher in c-Index performance \\
     \hline
\citep{schirris2022deepsmile} & colorectal and breast cancer& cancer subtype  & improves ROCAUC 3 to 10\% for HRD and MSI compared to the baseline\\
     \hline

\citep{naik2020deep} & breast cancer &molecular marker status  &AUC = 0.92 \\
     \hline
\citep{sandarenu2022survival} & breast cancer &Survival prediction   & 0.7179 in C-index\\
     \hline
\citep{hashimoto2022subtype} & malignant lymphoma & cancer subtype & ACC = 0.683\\
     \hline
\citep{butke2021end} & Urothelial Carcinoma & cancer classification  & ACC = 0.94\\
     \hline
\citep{yao2020whole} & Lung cancer & survival prediction & 0.606 in c-index\\
     \hline
\citep{zhao2021lung} & Lung cancer & cancer subtype  & AUC of 0.9602 in the ROI localization and an AUC of 0.9671 for subtype classification  \\
     \hline
\citep{wang2019rmdl} & Gastric cancer & cancer classification & ACC = 0.865\\
     \hline
\citep{su2022attention2majority} & regenerated kidney tissues &  cancer grading and multi-class problem   & ACC = 0.949, AUC = 0.974 \\
     \hline
\citep{del2021attention} & spitzoid melanocytic lesion diagnosis  &   cancer classification  & ACC = 0.9231\\
     \hline

\end{tabular} 
}
\end{table*}
\paragraph{Extend Attention-Based MIL approach to multi-head attention-based MIL}

The gated attention mechanism can also be extended from the binary classification of positive-negative based on the original MIL definition to multi-class classification. 
\citep{su2022attention2majority} apply the attention-based MIL approach with an intelligent sampling strategy into their weakly supervised framework on their in-house dataset to grade the regenerative kidney. They train a weakly supervised discriminator, apply the intelligent sampling method to the discriminator, and then extract features. Lastly, the MIL model is utilized to classify these intelligently sampled bags.
\citep{lu2021data} propose Clustering-constrained Attention Multiple instance learning (CLAM) with the attention mechanism and instance-level clustering to solve multi-class cancer classification problems.

Given $n$ class, the attention network will be split into $n$ parallel $ W_{1},W_{2},\cdots,W_{n}$, where $W = w^{T}$ (Eq.(\ref{equation:gated-attention})). For multi-class classification, the Eq.(\ref{equation:z}) and Eq.(\ref{equation:gated-attention}) is extended to:

\begin{equation}
z_{n} = \sum_{k = 1}^{K} a_{k,n}h_{k}
\end{equation}

\begin{equation}
a_{k,n} = \frac{ \exp { \{ W_{n}( \tanh  ( V_{n} h_{k}^{T}  ) \odot  sigm ( U_{n} h_{k}^{T} )) \}} }
{\sum_{j = 1}^{K} { \exp { \{ W_{n}( \tanh  ( V_{n} h_{j}^{T}  ) \odot  sigm ( U_{n} h_{j}^{T} )) \}} } }\qquad
\label{equation:gated-attention-multiclass}
\end{equation}
in which $a_{k,n}$ is the attention score for $ k^{th}$ instance for the $ n^{th}$ class, and the $z_{n}$ is the bag-level representation for the $ n^{th}$ class. In this way, different groups of instance feature vectors$ \{h_{1}, h_{2}, \cdots, h_{K} \} $ are learned. Here, $z_{n}$ is also known as the head, as the mechanism is known as multi-head attention.

Then, the head $z_{n}$ is utilized for predicting the bag-level (also called slide-level) score. For CLAM \citep{lu2021data}, the bag-level score is computed by separate fully connected layer as Eq.(\ref{equation:CLAM}):
\begin{equation}
s_{n} = W_{c,n} z_{n}^{T}
\label{equation:CLAM}
\end{equation}
For Attention2majority \citep{su2022attention2majority}, the bag-level score $s_{n}$ is given from the concatenation of multi-head $ z_{1}, \cdots, z_{n}$ as Eq.(\ref{equation:attention2majority}):
\begin{equation}
s_{n} = W_{O} concat (z_{1},\cdots, z_{n})^{T}
\label{equation:attention2majority}
\end{equation}

Finally, the softmax function is applied to bag-level prediction scores $s_{n}$ to make a final classification for the bag.

\subsubsection{CLAM}
\citep{lu2021data} first propose a novel interpretable model with the attention mechanism, which is called Clustering-constrained Attention Multiple instance learning (CLAM), to solve multi-class cancer classification problems. The attention network of CLAM returns scores for each patch and ranks those patches by their attention scores. The model selects the top-k most important and useless patches to learn the representative features. Those patches are the evidence of positive and negative. The model also visualizes the heatmap by attention scores to identify the region of interest and the patches with important cancer cell morphology for further diagnosis. Furthermore, CLAM is not limited to binary normal-tumor cancer classification and is also capable of cancer subtyping problems.
\citep{zeng2022artificial} apply CLAM architecture to develop models to predict the activation of 6 immune gene signatures. Their AI-based model is designed to extract biologically significant features from WSIs of hepatocellular carcinoma (HCC), and help improve immunotherapy in HCC patients. CLAM architecture generates attention maps, thus patches with high predictive value may contain important biological information. The morphological characteristics of these patches are consistent with the function of the genes included.

\subsubsection{DSMIL}
\citep{li2021dual} propose a dual-stream Multiple Instance Learning Network, which combines the instance-level and embedding-level MIL. The proposed dual-stream MIL Network utilizes a dual-stream architecture to learn the instance-level and the embedding-level classifier jointly. As we mentioned in the second part of this survey, instance-level, and embedding-level MIL have their advantages and disadvantages. One stream applies the instance-level MIL approach with max-pooling to identify the key instances. The second stream aggregates the instance into the embedding level to obtain further scores for the bag classifier. The final predictions are computed between key instances and each individual instance. Both instance-level and embedding-level scores are combined to produce the final score for the output bag score.

\subsubsection{Channel-Attention MIL}
\citep{wang2019rmdl} present a two-stage multiple instance learning framework for whole slide image classification of gastric cancer. This novel RMDL network selects patches with different contributions to predict the final label. In the first stage, channel attention is introduced into the model, which utilizes a localization network, to select the essential informative instances. In the second stage, the authors design an RMDL network to concatenate local-global features, so the model is not limited to local representation but also has the ability to capture global cell morphological information.

\subsubsection{Spatial-Attention MIL}
\citep{zhang2021joint} propose a novel spatial attention mechanism and sampling strategy based on magnification for cancer classification. The model computes the attention network at low magnification to generate the attention map, which represents the probability distribution of each patch, to select informative patches. Based on these potential informative patches, the model crops new patches at high magnification to predict the bag-level labels.

\subsubsection{Attention MIL with Hard negative mining}
One challenge of the MIL technique in histopathology is that the model incorrectly predicts negative instances as positive, due to imbalanced data and challenging instances. The imbalanced distribution of negative instances is caused by the property of WSI, as most regions of WSI are normal tissue, leading to the result that negative patches constitute a small proportion of the WSI. The model learns from limited negative instances together with a large number of positive instances. Additionally, difficult and challenging negative regions, referred to as hard negative patches, tend to be incorrectly classified as positive. Thus, \citep{li2019deep} propose their attention-based MIL approach with hard negative mining to address this challenge. They incorporate the attention mechanism and adaptive weighting strategy into the MIL model to learn instances' contribution to the final prediction. The attention scores of key instances enable the model to detect the hard negative instances leading to misclassification. Then, new hard negative bags with selected hard negative patches are generated for retraining to improve performance.

\citep{butke2021end} follow the approach proposed by \citep{li2019deep} to develop their novel method to classify the normal-cancerous urothelial cells, thus enabling the diagnosis of urothelial carcinoma. Given the dataset $X= \{X_{1}, X_{2}, \cdots, X_{N} \}$ with label $Y \in \{0,1\}$, and $H = \{h_{1}, h_{2}, \cdots, h_{K} \} $ be the embedding results. Different from \citep{li2019deep}, \citep{butke2021end} apply the Attention-based MIL pooling $z$ proposed by \citep{ilse2018attention} as Eq.(\ref{equation:z}) and the attention score $ a_{i}$ learned by Eq.(\ref{equation:gated-attention}) to their model. The learned attention scores are utilized to select hard negative instances. They select all bags $F_{L}$ of size $L$ from dataset $X$ with label $ Y = 0$ but are incorrectly predicted as positive, and then extract hard negative instances $M$ by:
\begin{equation}
M_{l} = \{a_{l,k}\mid a_{l,k} \ge \bar{a_{l}} + \frac{ \sigma_{l} } {5}   \}
\end{equation}
where $\bar{a_{l}}$ represents the average and $\sigma_{l} $ represents the standard deviation of attention scores for all false positive bag $F_{L}$. Then, each selected hard negative instance $M$ is passed through VGG16(pre-trained by ImageNet) as a feature extractor and a sampling stage to generate new hard negative bags.

\subsubsection{Other Attention MIL}
\citep{del2022constrained} firstly introduce constrained formulation into the MIL framework to predict ulcerative colitis activity based on neutrophil detection. The stringent criteria of Ulcerative colitis (UC) is the presence or absence of neutrophils. Their model incorporates location constraints to ascertain UC activity in whole slide images (WSI).

\citep{eastwood2023malignant} propose an end-to-end MIL approach with sampling-based strategy for subtyping malignant mesothelioma. This method employs an instance-based adaptive sampling scheme to train deep convolutional neural networks using bags of image patches. This approach enables learning from a broader range of relevant instances compared to traditional max- or top-N based MIL methods.

\subsection{Graph Neural Network}

Previous MIL methods extract features from selected regions of interest, which ignore the spatial connectivity of patches. Those methods cannot fully capture the tumors' continuous features and may cause Intratumoral Heterogeneity within the tumor.
Graph Neural Networks (GNNs) in whole slide image analysis provide a way to model the relationships between different regions of whole slide images. Whole slide images are often very large and high-resolution, and traditional CNNs can have difficulty processing all of the information in the image. GNNs allow the model to process the image in a more structured way, by modeling the relationships between different regions of the image as a graph.

In the whole slide image analysis, the original image is divided into non-overlapping patches, so some details and relationships between patches will be lost. GNNs can take advantage of the inherent structured data of whole slide image and are able to incorporate the context information of the regions of interest, which can improve the performance of the model.

\citep{wang2021hierarchical} propose a hierarchical graph-based deep learning framework to make a prediction of prostate cancer. They apply the Graph Convolutional Network with the attention mechanism to extract features along the graph, without the requirement of manually selecting ROIs and labels for each instance.
\citep{eastwood2023mesograph} propose a GNN-based MIL architecture to predict the subtype of mesothelioma, quantify the sarcomatoid component of tissue, generate association score as prognostic marker for survival prediction. They construct graph neural networks on tissue samples by treating each cell as an individual instance. This method allows for the exploration of differences in cell morphology across various subtypes. Additionally, it overcomes the constraints on spatial resolution of predictions that are typically imposed by a patch-based approach.

Each node in the graph is associated with a variety of features, which can be broadly categorized into four types. The proposed method represents each cell in a sample as a node within a graph, linking it to adjacent cells through edges that reflect spatial relationships and potential interactions.

\subsection{Vision Transformer}
\citep{shao2021transmil} introduce the transformer into MIL architecture to explore both morphological and spatial information. They propose a novel correlated MIL framework, named Transmil, to address the correlations among different instances with a single bag. Based on the correlated MIL framework, transformer is introduced to capture the spatial information and morphological information.

\citep{chen2022scaling} propose a hierarchical structure with visual transformer to learn features from both low-resolution and high-resolution. The 16 × 16 patches capture individual cells, and the 4096×4096 patches characterize interactions within the tissue microenvironment.

\citep{myronenko2021accounting} introduce self-attention transformer blocks and utilize an instance-wise loss function based on instance pseudo-labels to their MIL architecture to capture dependencies between instances. They achieve a good result for the Gleason grading task in prostate cancer.

\citep{jiang2024transformer} introduce ROAM, a novel region of interest (ROI)-based, multi-scale self-attention (SA) multiple instance learning (MIL) approach specifically designed to handle pathological images. This method leverages large-scale ROI analysis combined with self-attention mechanisms to extract detailed multi-scale information, which significantly enhances performance across a variety of glioma classification tasks. These tasks include tumor detection, subtype identification, grading, and prediction of molecular features. In addition, ROAM generates attention heatmaps at both the slide and patch levels. The interpretability of model greatly helps pathologists identify key diagnostic features, explore molecular features associated with cancers, and enhances overall understanding of cancers.

\subsection{ResNet-MIL}

\citep{campanella2019clinical} utilize ResNet34 to integrate instance-level feature representations to obtain final slide-level predictions (as shown in the figure. Their method, which names MIL-RNN, ranks the patches by their positive probabilities, studies these top-k higher-ranking patches, and passes them to RNN to make final predictions in order. They evaluate their model in 44732 WSIs without any expensive and labor-intensive annotations, achieving AUC greater than 0.98 in prostate cancer and basal cell carcinoma, and AUC greater than 0.96 in breast cancer metastases to axillary lymph nodes tasks.

\subsection{Pooling}
\citep{carmichael2022incorporating} introduce variance pooling into Attention-based multiple instance learning framework. \citep{ilse2018attention} to quantify intratumoral heterogeneity (ITH) on whole slide image. Cancer cells have different morphology within a tumor. Intratumor heterogeneity exists in all cancer types, due to the uncontrollable mutation and highly complex oncogenic signaling pathways of cancer. They also present interpretable results obtained from mean-pool and Var-pool. They observe that mean-pool capture patches contain dense and compact tumors, while var-pool capture fragmented tumors.

\subsection{Ehancements in Training Strategy}
\subsubsection{Pseudo label}
Given a dataset $X= \{X_{1}, X_{2}, \cdots, X_{N} \}$ of $N$ whole slide images with label $Y =  \{Y_{1}, Y_{2}, \cdots, Y_{N} \} \in \{ 0,1 \}$. Each slide $X_{i}$ is divided into non-overlapping patches $x= \{x_{i1}, x_{i2}, \cdots, x_{ij} \}$, constituting a bag $X_{i}$ with the bag-level label $Y_{i}$. The instances in each bag $X_{i}$ do not have instance-level labels. Models with pseudo label approaches select informative instances of size $k$ from patches $x= \{x_{i1}, x_{i2}, \cdots, x_{ij} \}$, and then create pseudo labels $y\in \{ 0,1 \}$ for selected informative instances. The selecting strategies can be top-k positive and negative attention scores, probabilities of instance, or sampled parameters of $\alpha $ \% positive and $\beta  $ \% negative patches.

\citep{qu2022dgmil} propose a feature distribution guided deep MIL framework (DGMIL) for Whole slide image analysis. They present that the feature distribution of whole slide images has a more significant improvement for cancer classification and the localization of positive patches. The pseudo label is introduced into their model to separate important positive and negative patches. The patches with the highest positive scores and the lowest positive scores are assigned to pseudo labels 1 or 0. These key instances with pseudo labels are put into the model to train the classifier. Finally, the Linear Projection Head is applied to map those key instance features into feature space.

\citep{lerousseau2020weakly} propose a weakly supervised MIL model, with a strategy that generates labels from whole slide images, for tissue-type segmentation tasks. According to the probability, $\alpha $ \% patches are assigned with label 1 and $\beta$ \% patches are assigned with label 0. Other patches are discarded while computing the loss. 
The pseudo label approach is also widely used in histopathology tasks. Moreover, other MIL approaches mentioned in this survey may also apply pseudo label mechanism as part of their approach.

\subsubsection{Student-Teacher distillation model}
Essentially, MIL models learn from WSIs and WSI-level labels to recognize the informative and distinctive patches. The general challenges of the WSI dataset in histopathology include the huge size of WSIs and the lack of annotation. Moreover, due to the inherent characteristic of the whole slide image, positive instances containing cancerous cells are only a small proportion of the dataset. All of these challenges lead to the overfitting problem of machine learning models. To address these challenges, recent works focus on the mutual relations between instances to reduce overfitting. However, these works are based on Attention-based MIL\citep{ilse2018attention}, which believes that it is infeasible to infer instance probabilities, and thus use attention score to represent the activation of positive instances. \citep{zhang2022dtfd} argue that attention score is not the strict criterion for the activation of positive instances. They infer the instance probability and introduce the pseudo bags into their double-tier feature distillation MIL framework (DTFD-MIL) based on the instance probability derivation. The proposed method randomly divides the original bags and assigns them labels from the original bag, with the risk of some pseudo bags with a label of 1 not containing any positive instances. Thus, the Tier-1 of the framework is utilized to estimate bag probabilities and train Tier-1 with its loss function. And then, the instance probabilities for each pseudo bag can be inferred, which are utilized to distill features. The distilled features are propagated to Tier-2 to generate better representations.

\subsubsection{End-to-End Training}
\label{end-to-end}

Most multiple instance learning approaches mentioned in the previous subsections are two-stage frameworks, which can be divided into two parts: a CNN encoder for extracting patches and an aggregation step for slide-level prediction. The CNN encoder and the aggregation module are decoupled in the prior two-stage framework. Consequently, the second stage of the model lacks control over the learned features, resulting in sub-optimal solutions \citep{sharma2021cluster}. Alternatively, the end-to-end framework, which uses a single model to learn features and then jointly optimize the whole model by one criterion, also provides a possibility for histopathology. It relies on weak labels assigned to the handcrafted region. However, the traditional end-to-end learning model is infeasible for whole slide image analysis due to the huge size of WSIs.

Attention-based MIL proposed by \citep{ilse2018attention} is a typical end-to-end architecture. In their study, the authors conduct an evaluation of their model using relatively small image sizes. Specifically, they utilize the BREAST CANCER dataset of 58 whole slide images (WSIs), each with dimensions of 896 by 768 pixels. It is noteworthy that other commonly used datasets in recent studies can be substantially larger to 150,000 by 150,000 pixels. It is impossible to train all patches directly in \citep{ilse2018attention}  work when the dataset is large. 

Similarly, \citep{das2018multiple} incorporate multiple instance pooling(MIP) into the MIL framework for end-to-end training. However, they still evaluate the performance on a small-scale dataset BreakHis \citep{spanhol2015dataset}, which comprises patches from regions of interest with the size 740 x 460 pixels rather than whole slide images. Additionally, the model only learns the feature representations with higher responses, which implies that a downsampling strategy must be employed to selectively choose certain parts of the instances from the whole set.

Therefore, \citep{sharma2021cluster} propose an end-to-end framework named Cluster-to-Conquer(C2C) with an adaptive attention mechanism to make the slide-level prediction, which enables to scale up to working with larger WSIs. Following feature extraction using ResNet-18, the authors employ the K-means algorithm to cluster all patches into $k$ clusters. From each of these clusters, $k'$ patches are sampled for further Attention-based aggregation, as Eq.\ref{equation:attention}. Finally, the aggregated representations are passed to the WSI classifier to make slide-level predictions. The C2C framework consistently generates clusters and assigns relevant patches with high attention weights.The entire model undergoes training via an end-to-end learning approach, employing an aggregated loss function. This function comprises both the KL-divergence loss, which is applied to the attention weights of patches originating from identical clusters, and the cross-entropy loss associated with both the whole slide images (WSIs) and individual patches.

The clustering strategy implemented in \citep{sharma2021cluster} is referred to as local clustering, which clusters patches from one single WSI as opposed to clustering patches from all WSIs. Different from the local clustering approach, \citep{xie2020beyond} propose an End-to-end Part Learning (EPL) framework, which involves end-to-end training with global clustering strategy that cluster patches from all WSIs.

\citep{chikontwe2020multiple} propose a novel MIL method with a center loss function that jointly learns both instance-level and embedding-level loss. In addition, they apply the top-k important instance selection strategy and soft-assignment based inference. The model computes the positive activation of each instance and generates instance-level probabilities. The Instance-level probabilities from all bags are then sorted in order to obtain top-k important instances of each bag for further training.

\section{Histopathological tasks}

MIL techniques utilized in computational pathology are primarily directed towards addressing two distinct categories of tasks~\citep{song2023artificial}. The first category contains basic tasks such as tumor detection, cancer grading, and subtype classification, collectively referred to as CPATH for automation. These tasks aim to automate the standard clinical workflow. Previous studies have demonstrated that these techniques can differentiate between benign and malignant cells with an accuracy comparable to that of experienced pathologists. The second category, often referred to as CPATH for discovery, involves advanced tasks that focus on predicting higher-level properties directly from whole slide images, with the goal of enhancing understanding of cancer characteristics and treatments. Certain genetic alterations are associated with identifiable specific morphological features. Recent studies have shown that these advanced computational techniques can accurately predict the mutation status of specific genes, as well as detect abnormalities in DNA repair mechanisms such as microsatellite instability and homologous recombination deficiency, using H\&E stained WSIs. In addition, these models are able to predict expression status of hormone receptor in breast cancer without the need for IHC tests targeting these receptors. In conclusion, the deployment of CPath has proven effective in identifing molecular features from H\&E WSIs.

\subsection{Basic Tasks in Computational Pathology for Automation}
In recent years, the field of computational pathology has witnessed a remarkable surge in clinical applications, driven largely by advancements in deep learning technologies. These innovations are designed to streamline complex workflows traditionally conducted by human pathologists. Key applications include the detection of tumor tissue in biopsy samples, cancer grading, and subtyping. A common feature of these deep learning approaches is their reliance on the same image data used by the ground-truth methods—typically, pathologists' assessments—for making predictions. For instance, the identification of invasive tumor tissue in prostate cancer biopsy samples is traditionally performed by pathologists examining H\&E-stained tissue slides. Correspondingly, a basic deep learning model trained on these H\&E histology images can perform the same task, predicting the presence of invasive cancer. Such foundational applications of deep learning serve to automate labor-intensive tasks, potentially reducing costs and decreasing turnaround times in pathology departments. However, they do not alter the fundamental nature of the diagnostic readouts used by clinicians to formulate treatment plans.

\subsubsection{Cancer classification}
Determining whether a tissue contains a tumor is the first step in routine diagnosis in cancer pathology. To enhance the accuracy and efficiency of this process, researchers have developed many noval deep learning methods capable of predicting tumor presence across various cancer types. ~\citep{qiu2021attention,butke2021end,wang2019rmdl,del2021attention} perform well for the task of binary or multi-class classification.

\subsubsection{Cancer grading}

The grading of cancerous tissues, which involves assessing the appearance of abnormal cells relative to healthy ones, is a critical component in the diagnosis of many diseases, including cancer. This grading process, performed by pathologists, relies on evaluating specific features such as glandular morphology and nuclear pleomorphism. For example, Gleason grading is the primary diagnostic and prognostic factor for prostate cancer. Traditionally, Gleason grading is carried out manually by expert pathologists based on H\&E tissue slides. However, MIL technique is capable of automating this task.
Many previous studies~\citep{zhang2021joint,myronenko2021accounting,li2021multi} with MIL approaches can now achieve Gleason grading performance that matches or exceeds that of pathologists.

The Gleason Grading System is the golden standard for the diagnosis and prognosis for prostate cancer. The current clinical practice first localizes the suspicious regions at low magnification and then utilizes these suspicious regions at high magnification for Gleason grading.
\citep{li2021multi} propose a  two-stage Attention-Based MIL architecture with a multi-resolution feature extract mechanism for prostate cancer classification and diagnosis on whole slide images. The first stage utilizes the attention network at low magnification to generate a saliency map, which represents the importance of each patch for predicting slide-level labels, to detect benign and malignant slides. The k-means algorithm is used for clustering. The second stage utilizes tiles selected from each k-means cluster at high magnification for prostate cancer grade classification.
\citep{myronenko2021accounting} introduce self-attention transformer blocks to their MIL architecture to capture dependencies between instances for the Gleason grading task.
\citep{zhang2021joint} propose a novel spatial attention mechanism and sampling strategy based on magnification for the classification of prostate cancer.

\subsubsection{Cancer Subtype}
Diagnostics in Computational Pathology (CPath) often involve subtyping, where the algorithm's goal is to classify cases into specific diagnostic categories. For instance, classifying lymph nodes based on whether they contain metastases, or distinguish between types of non-small cell lung cancer, such as adenocarcinoma versus squamous cell carcinoma~\citep{schirris2022deepsmile,hashimoto2022subtype,zhao2021lung}. These tasks are often important in clinic and required efficiently and accurately. Subtyping is utilized in diagnosing a range of diseases, including colorectal, skin, gastric, colon, liver, breast, and lung cancers. It also extends to identifying lymph node metastases and determining whether a tumor is primary or metastatic, as well as identifying its original site.

\subsection{Advanced Tasks in Computational Pathology for Discovery}

MIL techniques are approaching high accuracy in basic tasks in computational pathology. Despite these significant advances, many phenotypic information contained in WSIs remain underutilized for treatment decision making in basic tasks. Many studies have demonstrated that MIL can recognize patterns in high-dimensional data and directly predict the presence of a molecular feature, such as gene expression, mutations, hormone receptor status, or microsatellite instability from H\&E-stained WSIs. These molecular feature often cannot be reliably inferred by human experts from WSIs alone and require genomics assays such as IHC for further detection. Additionally, MIL has been applied to predict prognosis and treatment response in certain cancers. Unlike the basic tasks, these advanced tasks for the discovery of Cpath have the potential to revolutionarily impact clinical decision-making processes of cancers.

\subsubsection{Predicting Mutations}

Cancer is fundamentally a genetic disease, driven by oncogenic mutations~\citep{andor2016pan}. However, the mutation is uncontrollable and happens randomly in the human body. Oncogenic mutations represent a complex interplay between genetic and environmental factors that drive the transformation of normal cells into tumor cells. These mutations disrupt normal cellular functions and behaviors, leading to the development of cancer. As a result, genetic driver mutations can cause distinct morphological changes in cancer cells. For instance, BRAF mutations observed in melanoma are related to specific alterations in cell appearance, including larger size, rounder shapes, and increased pigmentation. \citep{zeng2022artificial} apply CLAM architecture to develop models to predict the activation of six immune gene signatures in hepatocellular carcinoma (HCC) directly from whole slide images, instead of molecular methods for extracting nucleic acid and RNA sequencing. Their AI-based model is designed to extract biologically significant features from WSIs of hepatocellular carcinoma (HCC) and help improve immunotherapy in HCC patients. CLAM architecture generates attention maps, thus patches with high predictive value may contain important biological information. The morphological characteristics of these patches are consistent with the function of the genes included. 
\citep{laleh2022benchmarking} research different supervised and weakly supervised approaches on the prediction of BRAF in colorectal cancer and FGFR3 mutation in bladder cancer. \citep{jin2024teacher} propose a Teacher-student MIL for the prediction of PDL1 expression for a broad range of cancers. Their approach utilizes MIL to predict PDL1 expression patterns and provide insights into histologies associated with PDL1. This demonstrates the potential of MIL in identifying various histological patterns indicative of molecular changes directly from H\&E images.

In addition, some types of cancers are often characterized by specific mutations. Cancers with specific mutations can be effectively treated with advanced treatments with lower mortality rates.The morphological features induced by single oncogenic driver mutations might be challenging for pathologists to identify directly and need comprehensive biomarker tests as an adjunct, but it is proved that deep learning can identify these patterns~\citep{echle2021deep,coudray2018classification}. Therefore, researchers have focused on detect genetic alterations directly from H\&E WSIs by deep learning. 

In lung cancer, EGFR mutations are present over 30\% of NSCLC in Asian patients~\citep{zhang2016prevalence}. This type of cancer can be benefited from targeted therapies~\citep{Mayo-nsclc}. Thus, detecting mutations directly from H\&E WSIs could have broad implications for clinical workflows.~\citep{zhao2022setmil} use multiple instances learning using a multiscale fusion module to predict the EGFR mutation in lung adenocarcinoma (TCGA-LUAD~\citep{tcga}), achieving an AUC of 0.838 in the gene mutation prediction task.~\citep{egfr} develop an attention-based MIL framework to improve EGFR predictive ability in lung adenocarcinoma. They also investigate regions with high attention scores from the MIL models to assess signal distribution. It shows that the MIL model has learned several known morphological and cytological associations with EGFR status from WSIs, achieving an AUC of 0.87.

Another clinically relevant example is about the BRAF mutation in thyroid cancer. BRAF is the driver gene that leads to different morphologies and affects treatment. 50\% of papillary subtype thyroid cancer and 10\% of follicular subtype thyroid cancer are associated with the BRAF V600E mutation and can be treated with targeted therapy.~\citep{anand2021weakly} utilizes their model to predict the BRAF mutation in thyroid cancer, and visualize informative regions to explore the different morphological features within a tumor.

\subsubsection{Predicting Gene Expression and Hormone Receptor Status}

\citep{naik2020deep} utilize their MIL model, named ReceptorNet, to determine ERS directly from H\&E staining whole slide images (WSI), bypassing IHC staining essayed by human experts. The tumor morphology captured in WSI is correlated with molecular biomarkers expressed in the IHC stain. The pathologist manually reviewed the high-attention patches with their ERs estimation. Furthermore, identifying discriminative tumor morphology for molecular biomarkers may deepen the biological understanding of how hormones drive tumor growth.
\subsubsection{Predicting Microsatellite Instability}

Microsatellite instability (MSI) is a genetic hypermutational state results from the defects of DNA mismatch repair proteins.
It is of great significance as a prognostic and predictive factor in clinical practice.
This biomarker can be a predictor of response to chemotherapy in colorectal cancer \citep{msib1}. It has the highest response rates to immunotherapy and has improved overall survival \citep{msib2}. 
Not all areas of the WSI provide MSI status information. Pathological indicators associated with MSI include the presence of mucin, poor or undifferentiated histology, and the presence of tumor-infiltrating lymphocytes (TILs). Due to intratumoral heterogeneity, it is unclear which tiles will contain these informative features without prior annotation. Therefore, assigning slide-level labels to each tile is imprecise and may introduce errors in training. One solution is weakly supervised learning, identifying MSI status. 
\citep{schirris2022deepsmile} apply their MIL model to downstream tasks of the Homologous recombination deficiency (HRD) classification in breast cancer and microsatellite instability (MSI) classification in colorectal cancer. \citep{laleh2022benchmarking} research different weakly supervised approaches on the prediction of microsatellite instability or mismatch repair deficiency from H\&E whole slide images.
\citep{cao2020development} develop an innovative Ensemble Patch Likelihood Aggregation model that incorporates a multiple instance learning (MIL) approach to predict microsatellite instability (MSI) status from whole slide images (WSIs). They implemented two distinct MIL pipelines to aggregate patch-level predictions into a comprehensive MSI score at the WSI level: the Patch Likelihood Histogram (PLH) pipeline and the Bag of Words (BoW) pipeline. To determine the most effective combination of these two MIL methods, they utilized ensemble learning techniques, which allowed them to optimize the convex combination of the methods and ultimately predict the MSI status of the patient accurately.

\subsubsection{Predict the Origin of the Primary Tumor}
 \citep{toad} propose an innovative attention-based algorithm designed to tackle two significant challenges in oncology: determining the origin of primary tumors and distinguishing between primary and metastatic tumors. These classification tasks are crucial because the site of the primary tumor significantly influences the management and treatment strategies for patients with metastatic cancers. Identifying the primary tumor's origin can often be complex, with histopathological examinations sometimes failing to provide clear answers.

\section{Discussion}

Multiple instance learning (MIL) is a machine learning technique that can be used to analyze whole slide images (WSIs). The goal of this survey is to investigate the potential of multiple instance learning (MIL) in whole slide image (WSI) analysis. The potential of MIL in WSI analysis includes:
\begin{enumerate}[i.]

    \item Discovery of cancer cell morphology: MIL can be used to identify and classify different types of cells within the whole slide images, which can assist pathologists in the discovery of new cell morphologies or variations within known cell types. 
    
    \item Interpretable machine learning models: MIL can be used to build interpretable machine learning models that can provide insights into the decision-making process of the cancer diagnosis, which can be useful in understanding the relationships between different regions of the whole slide image and the cancer grade.

    \item Processing large and high-resolution whole slide images: MIL can handle large, high-resolution images in specific medical tasks that are too complex for traditional machine learning methods to process effectively. MIL can also be utilized to train weakly labeled data, where only the labels of whole slide images are required, instead of instance-level labels and annotations of regions of interest.

\end{enumerate}

\subsection{Interpretable Machine Learning Models}
MIL approaches employed in WSI analysis necessitate a high degree of interpretability. Interpretability is essential for gaining human experts’ trust in automated diagnostic systems. Many machine learning models have been deployed into the black-boxes model. They are failing to explain how they learn the datasets to make predictions. 

In Multiple instance learning (MIL), interpretable machine learning models can be achieved by using attention mechanisms, pseudo-label, and instance-level attention, which can provide insights into the process of how the model works. For example, attention mechanisms can be used to identify the regions of interest in the original whole slide image. This explains which features of the image are most essential for the task, and where the model focuses on while making its predictions. The attention mechanism allows the model to provide the heatmap to show the important patches with different types of cancer cells or abnormalities, making it a powerful tool for pathology analysis and getting trust from doctors.

After histopathological tasks, such as cancer grading, cancer classification, and cancer subtype, the researchers also develop visualization tools to provide more interpretability for MIL models.

\citep{lu2021data} propose the interpretable model named CLAM to visualize the importance of informative regions.
They utilize the attention score to generate the heatmap. It is observed that CLAM can identify the boundary between normal and tumor tissue. In addition, high-attention regions correspond with well-known morphology identified by pathologists and human pathology understanding of tumors. The identified morphological characteristics are associated with diagnosis, prognosis, and response to treatment. The heatmap receives high Consistency with pathologist annotations, which means this visualization tool can provide highlight ROIs for pathologists in clinical trials.

\citep{anand2021weakly} present their model to predict BRAF mutation in thyroid cancer prediction. They also visualize the attention scores as a heatmap to highlight the determining regions and demonstrate interpretability (red for BRAF mutated regions and blue for wildtype regions). There is high spatial correspondence between highlighted regions obtained from the H\&E-based approach with IHC response regions from the IHC-based approach. These informative regions are enriched with morphologic features regarding BRAF mutation. Mutated tissue cells exhibit different morphological features with wildtype tissue.
Their interpretable model can help pathologists observe and explore the distinguishing features of tumors, earning the trust of pathologists and physicians. In addition, the attention map can guide pathologists to target the regions for further molecular methods of mutation testing. 

\citep{zeng2022artificial} utilize their MIL model to predict the activation of six immune gene signatures in hepatocellular carcinoma (HCC) directly from whole slide images. Patients with these immune genes can benefit from particular treatment options - immunotherapy. They apply CLAM approach to generate a heatmap and explain how each tissue sections contribute to the machine learning model to determine the region enriched with informative features. The informative patches captured by models are further analyzed by pathologists for morphological features of hepatocellular carcinoma (HCC). They demonstrate that the informative patches regarding interferon gamma signaling are highly associated with lymphocytes, neutrophils, and plasma cells, which are consistent with the functions of the gene. This interpretable machine learning model provides insight into the biological understanding of hepatocellular carcinoma (HCC).

\subsection{Discovery of Cancer Cell Morphology}

Genetic driver mutations can change the morphology of cancer cells. However, the vast majority of machine learning pipeline focus on associating tissue morphology with the expression level of individual genes. The input is a WSI and the output label is the expression level of a single gene. However, the reality is that cancer is the correlated expression of multiple genes. Discovering new cell morphology is a complex task that requires understanding cell biology, pathology, and the ability to interpret various results from tissues and cells. Deep learning can be used as a tool to assist pathologists in the analysis of WSIs, by providing a fast and efficient way to identify the new cancer cell morphology within WSIs. 
\citep{carmichael2022incorporating} utilize their MIL model with variance pool to investigate intratumoral heterogeneity of cancer. The mean-pool and var-pool result on patches with different cancer cell morphology. 
\citep{zeng2022artificial} utilize their MIL model to predict the activation of six immune gene signatures in hepatocellular carcinoma (HCC) directly from whole slide images, instead of molecular methods to extract nucleic acid and RNA sequencing. The generated heatmaps enable them to identify informative patches and investigate the different morphological characteristics of hepatocellular carcinoma (HCC).
\begin{table*}[ht]
	\caption{satisfactory performing model for different train dataset size}\label{tab:table3}
 	\resizebox{\textwidth}{!}{
	\begin{tabular}{ p{3cm} | p{6cm} | p{5cm} | l |l | p{5.5cm} }
		\hline		
Papers& model & datasets & \#training WSIs & AUC & tasks \\ 
  \hline
\citep{lu2021data} & CLAM &\multirow{2}*{TCGA-RCC} &25\% train data(170 WSIs) &0.9532& RCC subtyping task \\
& &  &100\% train data &0.9915& \\
     \hline   
\citep{lu2021data} & CLAM & \multirow{2}*{TCGA-NSCLC + CPTAC-NSCLC} &25\% train data(419 WSIs)  &0.9032& NSCLC subtyping task \\
     & &  &100\% train data &0.9561& \\
     \hline   
\citep{lu2021data} & CLAM & Camelyon16 +  Camelyon17 \newline\citep{c17}&50\% train data(289 WSIs) &0.9286& Lymph node metastasis detection task \\
& &  &100\% train data &0.9532& \\
     \hline  
\citep{schirris2022deepsmile} & DeepSMILE(SimCLR-VarMIL) &\multirow{3}*{TCGA-BC }&40\% train data &0.77& HRD classification task \\
& DeepSMILE(SimCLR-VarMIL)&  &100\% train data &0.81& \\
& DeepSMILE(ImageNet tile-supervision)&  &100\% train data &0.77& \\
     \hline
\citep{schirris2022deepsmile} & DeepSMILE(SimCLR-VarMIL) &\multirow{4}*{TCGA-CRC} &40\% train data &0.75& MSI prediction task \\
      & DeepSMILE(SimCLR-VarMIL)&  &60\% train data &0.80& \\
   & DeepSMILE(SimCLR-VarMIL)&  &100\% train data &0.86& \\
& DeepSMILE(ImageNet tile-supervision)&  &100\% train data &0.77& \\
     \hline  
\end{tabular} 
}
\end{table*}

\subsection{Simplify the Workflow}

MIL approaches have been widely used in different histopathological tasks for WSI analysis. Common applications of such techniques include tumor detection, cancer subtyping, and grading. MIL approaches have also been employed to the tasks that simplify the current workflow and surpassing human capabilities. Such tasks include the prediction of molecular alterations, survival outcomes, and the activation of driver genes directly from original WSIs. These broad applications in WSI analysis are expected to be widely adopted in clinical practice. H\&E-WSIs are available for almost every cancer patient, especially for patients from low-income areas. Hence, MIL tenique are expected to integration into existing diagnostic pathways,potentially improving patient outcomes and reducing costs. IHC is one of the examples.

According to the guidelines of the US National Comprehensive Cancer Network, Estrogen receptor status (ERS) is a crucial criterion for the diagnosis and prognosis of breast cancer, and the detection of ERS is essential for every new patient. Pathologists determine ERS by visually inspecting the results of immunohistochemistry (IHC) staining with an antibody targeting the ER receptor on tissue samples. 

Despite its widespread use in medical research and diagnostics, immunohistochemistry (IHC) has several limitations. The IHC process is expensive, time-consuming, and labor-intensive. This will be a huge concern for low-income countries. In addition, IHC relies more on the staining protocol, in which antibodies are only available for the most common mutation (e.g., BRAF V600E), and the output is expressed by the intensity of the stain, the percentage of cells with detectable staining, or the presence/absence of a stain. In addition, there is variability in IHC preparation due to the different antibody resources and clones, technician skill levels, and tissue handling procedures. Moreover, these human visual inspections are inherently subjective, and human errors may occur during the decision-making process.

Another clinically relevant example is about BRAF mutation in thyroid cancer. BRAF is the driver gene that leads to different morphologies and affects treatment. 50\% of papillary subtype thyroid cancer and 10\% of follicular subtype thyroid cancer are associated with the BRAF V600E mutation and can be treated with targeted therapy. Detection of BRAF mutation by using machine learning models contributes significantly to thyroid cancer prognosis and treatment, as IHC relies more on staining protocols, antibodies are only available for the most common mutation(e.g. BRAF V600E), and the cost of DNA analysis is a huge concern for low-income countries.~\citep{anand2021weakly} utilize their model to predict BRAF mutation in thyroid cancer prediction. Moreover, their visualizations of informative regions explore the different morphological features within a tumor and provide a way to deepen the biological understanding of cancers.

\subsection{Rare Disease}

\citep{campanella2019clinical} claims that at least 10000 WSIs are needed for training to achieve good performance.

For rare diseases (e.g., Chromophobe Renal Cell Carcinoma), thousands of labeled WSIs are difficult and even impractical to collect or acquire. \citep{lu2021data} claim that model CLAM achieves satisfactory performance on small datasets for some rare diseases. For three-class cancer subtyping of Renal Cell Carcinoma (RCC), 170 WSIs for each class is enough to achieve a test of AUC \textgreater 0.94. The required amount of WSIs varies depending on the different cancer classification. For Non-small Cell Lung Carcinoma (NSCLC) subtyping and lymph node metastasis, 419 WSIs and 289 WSIs are needed.

\citep{schirris2022deepsmile} claim that their model can reduce over 50\% labeled training data. For the HRD classification task in TCGA-BC by using DeepSMILE (SimCLR-VarMIL), 40\% training data achieve a higher AUC than 100\% training data with the ImageNet tile-supervision approach.

\section{Conclusion}
In this survey, we present a comprehensive overview of multiple instance learning models have been developed in histopathology analysis. We summarize the methodological aspect of different multiple instance learning approaches such as attention mechanism, pseudo label, transformer, contrastive learning, and end-to-end training. We also provide an overview of multiple instance learning approaches applicable to different diseases and histopathological tasks. Finally, we have outlined the potential of multiple instance learning and future trends for the progress of CPath.

\bibliographystyle{plainnat}
\bibliography{ref} 

\end{document}